\pdfoutput=1

\documentclass[11pt]{article}

\usepackage{arabtex}
\usepackage{utf8}
\usepackage[LAE,LFE]{fontenc}

\usepackage{longtable}

\usepackage[table]{xcolor}
\usepackage{colortbl} 

\usepackage{pifont}
\usepackage{amssymb}
\usepackage{amsthm}

\newcommand{\cmark}{\textcolor{teal}{\ding{51}}}
\newcommand{\xmark}{\textcolor{red}{\ding{55}}}

\usepackage[preprint]{acl}

\usepackage{times}
\usepackage{latexsym} 

\usepackage[T1]{fontenc}

\usepackage[utf8]{inputenc}

\usepackage{microtype}

\usepackage{inconsolata}

\usepackage{graphicx}
\usepackage{multirow}
\usepackage{subcaption}
\usepackage{amsmath} 
\usepackage{enumitem}
\usepackage{booktabs}

\usepackage[utf8]{inputenc}

\newtheorem{hypothesis}{Hypothesis}

\newcolumntype{R}[1]{>{\raggedleft\arraybackslash}p{#1}}

\newcommand{\ar}[1]{\setcode{utf8}\RL{#1}}
\newcommand{\emoji}[1]{\includegraphics[height=2.5ex]{#1}}

\title{AL-QASIDA: Analyzing LLM Quality and Accuracy Systematically in Dialectal Arabic}

\author{
Nathaniel R. Robinson$^{2}$ 
\quad Shahd Abdelmoneim$^{1}$ 
\quad Kelly Marchisio$^{1}$
\quad Sebastian Ruder$^{1}$
\\
$^1$Cohere,
$^2$Johns Hopkins University\\
 \texttt{nrobin38@jhu.edu, kelly@cohere.com}\\
}

\begin{document}
\maketitle
\begin{abstract}
Dialectal Arabic (DA) varieties are under-served by language technologies, particularly large language models (LLMs). 
This trend threatens to exacerbate existing social inequalities and limits LLM applications, yet the research community lacks operationalized performance measurements in DA. 
We present a framework that comprehensively assesses LLMs' DA modeling capabilities across four dimensions: fidelity, understanding, quality, and diglossia. 
We evaluate nine LLMs in eight DA varieties and provide practical recommendations. 
Our evaluation suggests that LLMs do not produce DA as well as they understand it, not because their DA fluency is poor, but because they are reluctant to generate DA. 
Further analysis suggests that current post-training can contribute to bias against DA, that few-shot examples can overcome this deficiency, and that otherwise no measurable features of input text correlate well with LLM DA performance. 

\end{abstract}

\section{Introduction} 
\label{sec:intro}


Large language models (LLMs) have transformed natural language processing (NLP) for many languages \cite{singh-etal-2024-aya,dubey2024llama3herdmodels}. 
However these technologies often lack support for minority dialects and language varieties \cite{robinson-etal-2023-chatgpt,zhu-etal-2024-multilingual,arid2024large,joshi2024natural,chifu-etal-2024-vardial}.
Arabic, the fourth most spoken macro-language in the world with 420M speakers and official status in 26 countries \cite{bergman-diab-2022-towards,tepich2024influence}, has a diversity of such varieties.  
Despite its prominence, Arabic has been historically set back in NLP due to its script, which lacked digital support until the 1980s \cite{parhami2019evolutionary}. 

\begin{figure}
    \centering
    \includegraphics[width=1\linewidth]{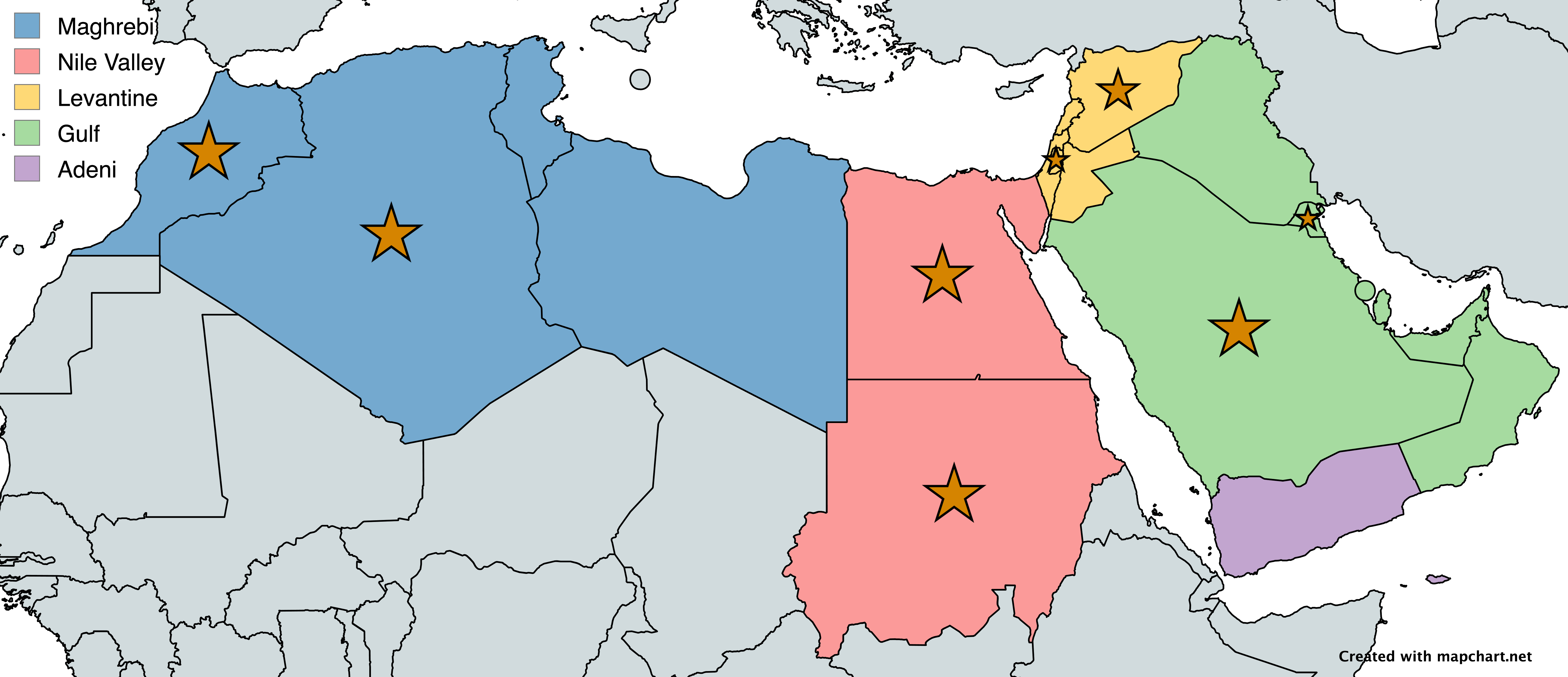}
    \caption{Arabic greater dialectal regions, per \citet{diab-habash-2007-arabic}. Stars indicate the eight nations whose DA varieties are represented in this work.}
    \label{fig:dial_reg}
\end{figure}

\begin{table*}
  \centering
  \small 
  \begin{tabular}{lrr}
    MSA     & \ar{أريد أن أخبرك بشيء جيد جداً}& \ar{لم أرَ كوب الماء هذا}\\
    Egyptian DA    & \ar{عايز اقلك حاجة كويسة أوي} & \ar{مشفتش كباية المية دي}\\
    Jordanian DA     & \ar{بدي قولك اشي كتير منيح} & \ar{ما شفت هاي الكاسة للماي}          \\
    \hline
    English     & I want to tell you something very good & I didn't see that glass of water           \\
  \end{tabular}
  \caption{Example Arabic sentences with 0\% word overlap across three varieties}
  \label{tab:zero_overlap}
\end{table*}

Many NLP tools support Arabic now, but often only Modern Standard Arabic (MSA); the numerous Dialectal Arabic (DA) varieties are often neglected. 
There are 28 microlanguages designated as Arabic in ISO 639-3,\footnote{\url{https://wikipedia.org/wiki/ISO_639_macrolanguage}} of which MSA is only one. 
In the Arab world, MSA is used only in narrow circumstances, while local DA varieties are predominant \cite{bergman-diab-2022-towards,barz-jstor}. 
MSA has no native speakers according to Ethnologue;\footnote{\url{https://www.ethnologue.com/}} Arabic speakers speak their DA varieties natively (as L1) and later learn MSA as an L2 \cite{azzoug2010modern}. 
Many who lack educational resources are not proficient in MSA \cite{bergman-diab-2022-towards}. 
Arabic varieties are diverse and differ both phonologically, morphologically, syntactically, semantically, and lexically \cite{habash2010introduction,keleg-etal-2023-aldi}. According to \citet{bergman-diab-2022-towards}, Moroccan Arabic or Darija (\texttt{ary}) and Egyptian Arabic (\texttt{arz}) are as mutually intelligible as Spanish and Romanian. 
Table \ref{tab:zero_overlap} illustrates two simple sentences that display 0\% word overlap across three Arabic varieties. 
\citet{bergman-diab-2022-towards} have urged researchers to move beyond treating Arabic as a "monolith," i.e. aiming only for MSA support.

Many LLMs today are proficient in MSA but reluctant to model DA; users wishing to converse in DA often have to perform extensive prompt acrobatics to coerce the LLM to use the dialect.\footnote{For instance, when conversing with ChatGPT in Egyptian Arabic, it took one of the authors 14 turns of conversation before the LLM used the desired dialect (see Table \ref{tab:gpt_convo}).} 
Because MSA proficiency correlates with educational settings (and as a result socioeconomic advantage), such disparity may exacerbate existing inequalities.
Even when interacting with MSA-proficient users, replying to informal DA inputs with formal MSA sounds unnatural and limits an LLM's uses. 

LLM pre-training data is diverse and likely includes content in many Arabic dialects. 
However, post-training uses existing human-labeled data, which is typically MSA, and newly collected data that commonly adopt an official tone corresponding to a formal MSA register. 
Current LLM behavior suggests that models have the capability to understand and model DA, but that they default to MSA regardless, which could be caused by biases introduced in post-training. 
We thus hypothesize:

\begin{hypothesis}\label{hyp:sft_bad}
Current post-training methods make LLMs more reluctant to model DA.
\end{hypothesis}

\begin{hypothesis}\label{hyp:understanding}
Post-trained LLMs understand DA better than they generate it.
\end{hypothesis}



Arabic-speaking communities are aware of LLMs' DA shortcomings, but lack a standard operationalized definition of LLM DA proficiency. 
To this end, we present an evaluation suite of DA proficiency along different dimensions. 
We contribute:
\setlist{nolistsep}
\begin{enumerate}[nolistsep]
    \item AL-QASIDA: a method to evaluate LLM DA proficiency along dimensions of \textbf{fidelity}, \textbf{understanding}, \textbf{quality}, and \textbf{diglossia}.
    \item Best practice recommendations, including use of few-shot prompts for DA generation, Llama-3 for monolingual tasks, and GPT-4o for cross-lingual requests of Egyptian or Moroccan varieties.
    \item The following six key findings from our AL-QASIDA evaluation and analysis.
\end{enumerate}

\noindent
These findings include:
\setlist{nolistsep}
\begin{enumerate}[nolistsep]
    \item \textbf{LLMs do not produce DA as well as they understand it} (as Hyp. \ref{hyp:understanding}). This is an apparent reversal of the Generative AI Paradox \cite{west2024generative}, which observes that models' generative capabilities exceed their ability to understand the generated output.
    \item When LLMs do produce DA, they do so without perceptible declines in fluency. 
    \item LLMs are not diglossic: they generally cannot translate well between MSA and DA. 
\end{enumerate}

\noindent
And further analysis suggests the following: 
\setlist{nolistsep}
\begin{enumerate}[nolistsep]
    \item Post-training can bias LLMs against DA (see Hyp. \ref{hyp:sft_bad}), but otherwise improves text quality. 
    \item Few-shot prompting improves DA proficiency across dialects and genres. 
    \item Otherwise, no input text features correlate strongly with LLM DA performance.  
\end{enumerate}

\section{Related Work and Background}

There exist notable prior works on benchmarking NLP for DA varieties. 
AraBench \cite{sajjad-etal-2020-arabench} is a benchmark for MT between Arabic varieties and English that predates LLMs' widespread popularity. 
DialectBench \cite{faisal2024dialectbench} is a benchmark of 10 text-based traditional NLP tasks---such as parsing, part-of-speech tagging, and named entity recognition---that covers a large number of dialects and varieties in various languages, including multiple DA varieties. 
ARGEN \cite{nagoudi-etal-2022-arat5} compares mT5 \cite{xue-etal-2021-mt5} with novel AraT5 de-noising language models across MT, summarization, paraphrasing, and question generation tasks. 
Dolphin \cite{nagoudi-etal-2023-dolphin} is a comprehensive Arabic NLP benchmark that evaluates DA, but does not define DA proficiency.
AraDICE \cite{mousi2024aradice} is an LLM benchmark much like ours that focuses on accuracy and cultural appropriateness in DA. 
Our work differs from these in its purpose: to define LLM DA proficiency. 
To our knowledge, ours is the first evaluation to measure this comprehensively.

\subsection{Arabic Dialect Identification}
\label{sec:nadi}


\begin{table*}[]
    \centering
    \small 
    \begin{tabular}{rcccccr}
         & \emoji{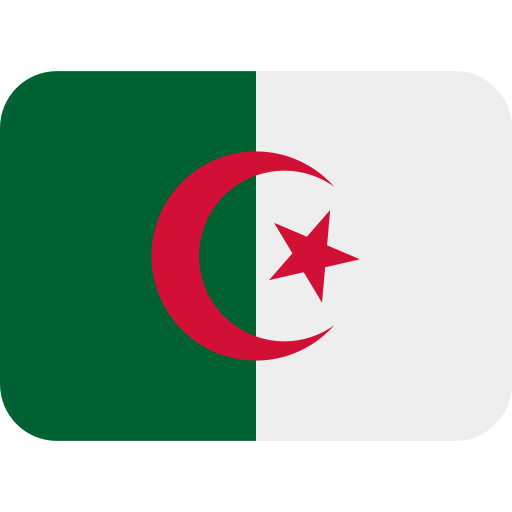} & \emoji{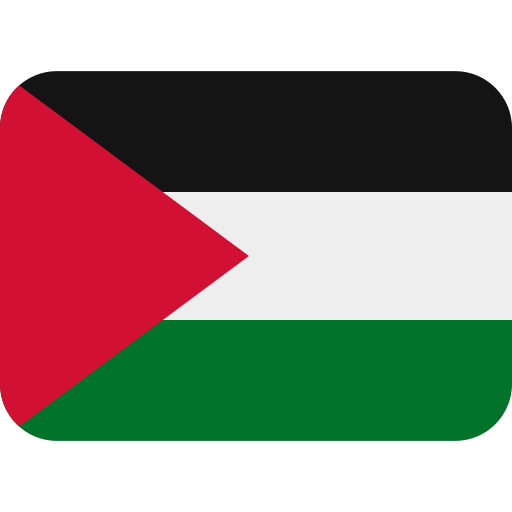} & ... & \emoji{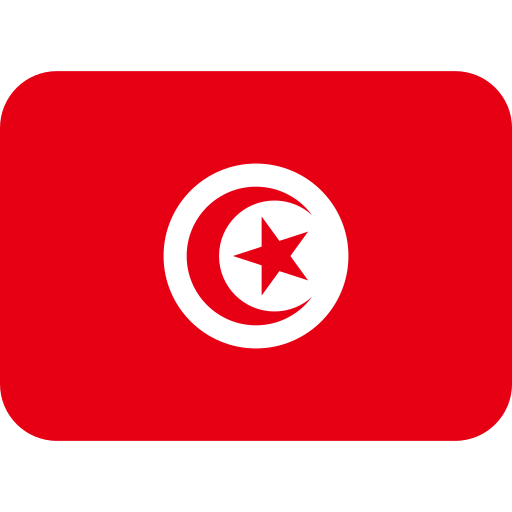} & \emoji{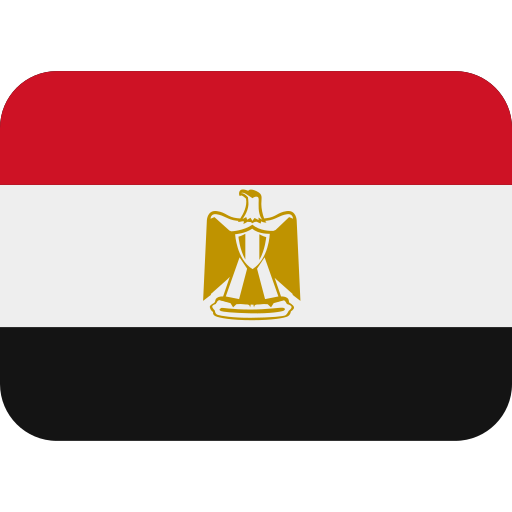} & \textit{ALDi} \\
         \hline 
         \ar{يا ريتني ما تفرجت عالحلقة الأخيرة تاعت رحيم تأزمت بصراحة أبدع ياسر جلال ...}& \cmark & \xmark & ... & \cmark & \xmark & High (1) \\
         \ar{هوا ورياح الشبابيك والابواب بتخبط كانه زلزال ربنا يستر}& \xmark & \cmark & ... & \xmark & \cmark & Med. (.7)
    \end{tabular}
    \caption{One sentence is valid in Algerian and Tunisian DA, while another is valid in Palestinian and Egyptian, with varying ALDi (dialectness). Adapted from \href{https://nadi.dlnlp.ai}{\texttt{nadi.dlnlp.ai}}. See also Table 1 of \citet{abdul-mageed-etal-2024-nadi}.}
    \label{tab:multi_label}
\end{table*}

\begin{figure}
    \centering
    \includegraphics[width=1\linewidth]{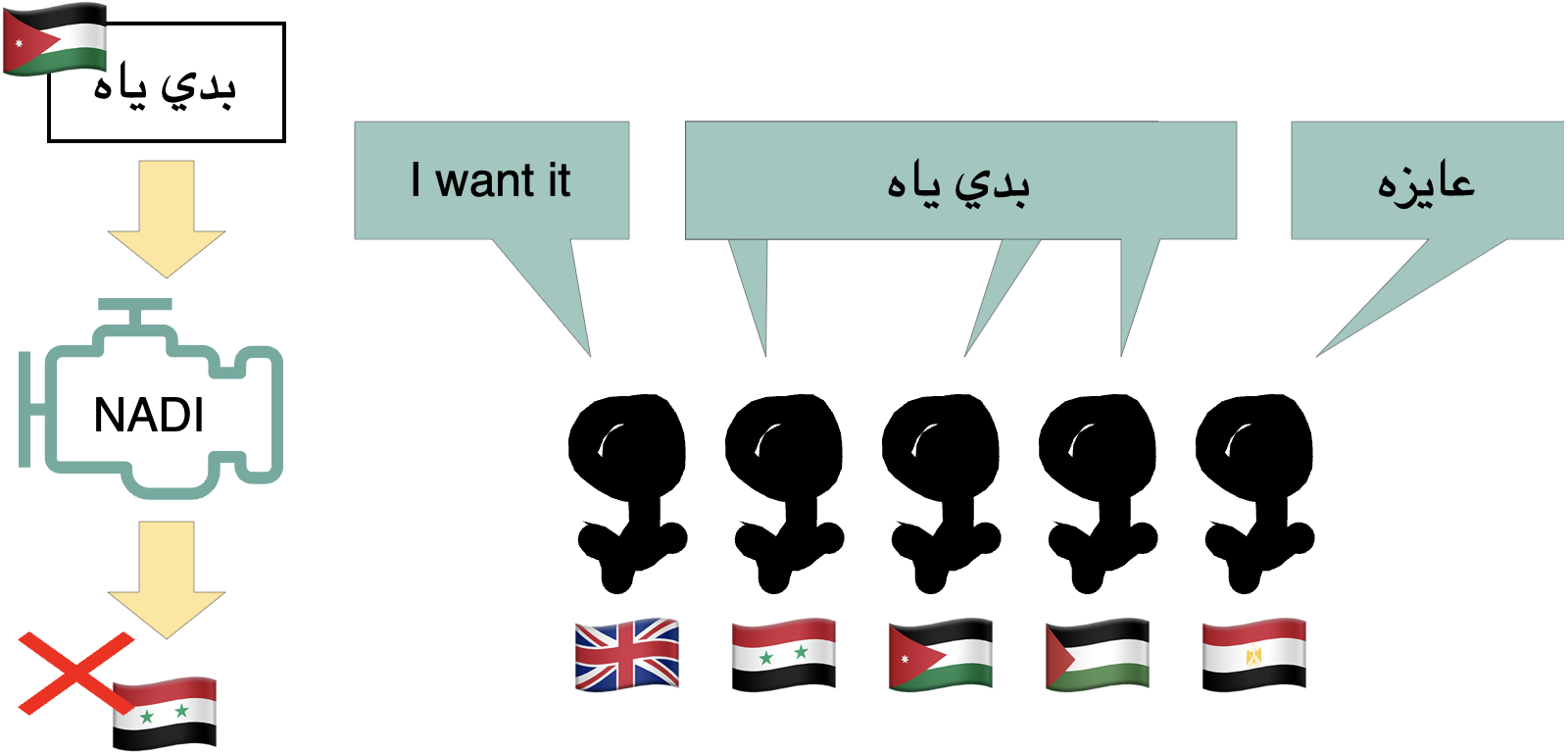}
    \caption{A sentence shared across Syrian, Jordanian, and Palestinian varieties may be labeled as Jordanian but predicted as Syrian, resulting in a false NADI error.}
    \label{fig:stick_figs}
\end{figure}

Identifying Arabic varieties, called Nuanced Arabic Dialect Identification (NADI), has been researched for years. 
Until recently, NADI shared tasks \cite{bouamor-etal-2019-madar} evaluated models that produce a single country-level or city-level dialect label for each input sentence. 
However, this approach was found insufficient for DA intricacies. 
Arabic varieties have significant overlap, especially in text for geographically proximate varieties \cite{abdul-mageed-etal-2024-nadi}. 
For example, Figure \ref{fig:stick_figs} illustrates a sentence that is valid in multiple Levantine countries.  
If a NADI model labeled such a sentence as Syrian while the ground-truth label were Jordanian, this would be falsely deemed an error. 
In fact, \citet{keleg-etal-2023-aldi} found that 66\% of a single-label NADI model's supposed errors were not errors at all. 
Hence the most recent NADI shared task \cite{abdul-mageed-etal-2024-nadi} focused on multi-label NADI: mapping an input sentence to a multi-hot vector label, as in Table \ref{tab:multi_label}. 

\citet{keleg-etal-2023-aldi} marked another change in standard NADI approach. 
Instead of treating MSA as an additional variety alongside DA varieties \cite{salameh-etal-2018-fine}, \citeauthor{keleg-etal-2023-aldi} framed MSA identification as a separate task: Arabic Language Dialectness (ALDi). 
ALDi models predict where a given utterance falls on the dialectness scale, with a $\mathrm{score}_{\mathrm{ALDi}}$ of 0 being fully MSA and 1 being fully DA (regardless which DA variety). 
See Table \ref{tab:multi_label}.

\section{Methodology}
\label{sec:meth}

To be proficient in DA, an LLM requires different competencies that we define as the following:
\setlist{nolistsep}
\begin{enumerate}[nolistsep]
    \item \textbf{Fidelity:} Can the LLM identify and produce the correct DA variety when asked? 
    \item \textbf{Understanding:} Does the LLM understand prompts in the DA variety? 
    \item \textbf{Quality:} Is the LLM able to model the DA variety well (i.e., does its quality deteriorate compared to MSA or another language)?
    \item \textbf{Diglossia:} Can the LLM translate between DA and MSA?
\end{enumerate}


\textbf{Fidelity} is crucial as it forms the prerequisite for further assessment. \textbf{Understanding} of DA prompts and \textbf{Quality} of DA responses are both necessary for successful user interactions. Finally, \textbf{Diglossia} measures whether an LLM is aware of fine-grained differences between DA and MSA.

\subsection{Operationalizing Fidelity}
\label{sec:fidelity}



\begin{table}[]
\resizebox{\columnwidth}{!}{%
\begin{tabular}{lllll}
\textbf{Dimension} & \textbf{Capability} & \begin{tabular}[c]{@{}l@{}}\textbf{Input}\\ \textbf{lang.}\end{tabular} & \begin{tabular}[c]{@{}l@{}}\textbf{Output}\\ \textbf{lang.}\end{tabular} & \textbf{Metric} \\ \midrule
\multirow{2}{*}{Fidelity} & \begin{tabular}[c]{@{}l@{}}Monolingual\\ generation\end{tabular} & DA & DA & ADI2 score \\
 & \begin{tabular}[c]{@{}l@{}}Cross-lingual\\ generation\end{tabular} & eng & DA & ADI2 score \\ \hline
\multirow{2}{*}{Understanding} & Translation & DA & eng & spBLEU \\
 & \begin{tabular}[c]{@{}l@{}}Instruction\\ following\end{tabular} & DA & DA & Human eval \\ \hline
\multirow{2}{*}{Quality} & Translation & eng & DA & spBLEU \\
 & Fluency & DA/eng & DA & Human eval \\ \hline
\multirow{2}{*}{Diglossia} & Translation & MSA & DA & spBLEU \\
 & Translation & DA & MSA & spBLEU \\ 
\end{tabular}%
}
\caption{Evaluation data and metrics for the four competencies to assess DA proficiency in LLMs. 
Data are in dialectal Arabic (DA), Modern Standard Arabic (MSA), or English (eng).
}
\label{tab:competencies}
\end{table}

To measure how well the LLM can identify and produce requested DA varieties (\textbf{Fidelity}), we evaluate whether the LLM produces the desired DA variety in monolingual (prompting the LLM in a specific DA variety) and cross-lingual (requesting a specific DA variety from the LLM in English) settings, analogous to \citet{marchisio2024understanding}.

To this end, we require a NADI model to identify the Arabic variety of LLM outputs. 
As single-label NADI classifications can lead to false negatives in evaluation \cite{abdul-mageed-etal-2024-nadi}, as seen in Figure \ref{fig:stick_figs}. 
To mitigate this, we apply the NADI 2024 shared task baseline model\footnote{The model was originally trained for the single-label task of \citet{abdul-mageed-etal-2023-nadi}.} \cite{abdul-mageed-etal-2024-nadi}, but extract the probability of the desired dialect from the output logits rather than using the one-hot classification output.
The NADI score ($\mathrm{score}_\mathrm{NADI}(\cdot)_C$) of an LLM is then the probability that its output is in the desired DA variety of country $C$.
As NADI and ALDi are independent (see \S \ref{sec:nadi}), this NADI model does not distinguish between MSA and DA: only between country-level varieties. 
We therefore employ an additional model for ALDi: to differentiate MSA and DA \cite{keleg-etal-2023-aldi}. 
We finally define a DA fidelity performance metric that combines both NADI and ALDi scores: Arabic Dialect Identification And DIalectness (ADI2) score.
Given LLM output $y$,\footnote{Because some LLMs, especially base models, tend to repeat DA inputs in monolingual settings, we remove any copies of the prompt from output $y$ before ADI2 computation.} we define:

\begin{figure*}
    \centering
    \includegraphics[width=\linewidth]{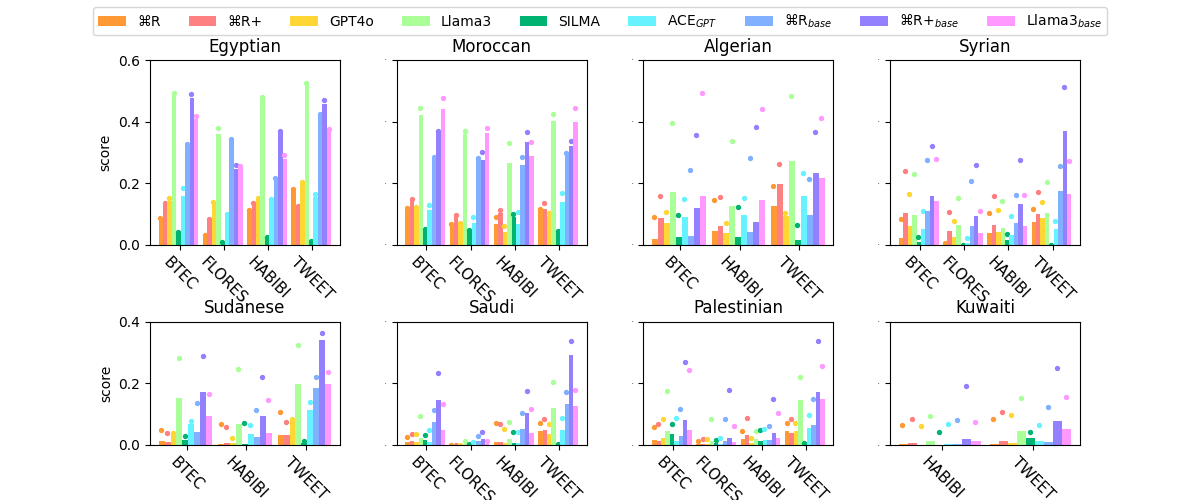}
    \caption{Llama models and Command series base models are best at maintaining the user's DA variety, as measured by ADI2 score (bars) and macro-score (marks).
    }
    \label{fig:mono-res}
\end{figure*}



\begin{align*}
\mathrm{score}_\mathrm{ADI2}(y) & = P(y \text{~is~dialectal~} C \text{~Arabic}) \\
& = P(y \text{~is~DA,~} y \text{~is~} C \text{~Arabic}) \\
& = P(y \text{~is~DA}) * P(y \text{~is~} C \text{~Arabic}) \\
& = \mathrm{score}_\mathrm{ALDi}(y) * \mathrm{score}_\mathrm{NADI}(y)_C \\
\end{align*}


Because some DA varieties are similar (see \S \ref{sec:nadi}), we compute an alternative to $\mathrm{score}_\mathrm{NADI}(\cdot)_C$, which we call the NADI macro-score: $\mathrm{macro}_\mathrm{NADI}(y)_R := \sum_{C \in R} \mathrm{score}_\mathrm{NADI}(y)_C$, where $R \in \{ \text{\small \textbf{Maghreb}},~\text{\small \textbf{Nile Valley}},~\text{\small \textbf{Levant}},~\text{\small \textbf{Gulf}} \}$, with each region $R$ itself representing a set of countries $C$. 
Accordingly, $\mathrm{macro}_\mathrm{ADI2}(y)_C := \mathrm{score}_\mathrm{ALDi}(y) * \mathrm{macro}_\mathrm{NADI}(y)_R$, given $C \in R$. 
This metric indicates whether the LLM responds with DA varieties from the correct region, and thus allows for NADI confusions between proximate varieties. 

\subsection{Operationalizing Other Competencies}

\paragraph{Understanding} 

We use DA-to-English translation to evaluate the LLM's understanding of DA text. 
As English is the closest language to an LLM's internal representation \cite{etxaniz-etal-2024-multilingual}, we can assess to what extent the model understood the original DA text based on the quality of its English translation.
Machine translation (MT) metrics such as BLEU \cite{papineni-etal-2002-bleu} or human annotations thus serve as proxies to DA understanding. 
We employ SpBLEU \cite{goyal-etal-2022-flores} and chrF \cite{popovic-2015-chrf}. 
In addition to assessing \textbf{Understanding} via MT, we ask human annotators to evaluate understanding directly by determining to what extent the LLM fulfilled requests in monolingual DA prompts during the \textbf{Fidelity} evaluation.\footnote{We rely on humans as LLM-as-a-judge evaluators \cite{NEURIPS2023_91f18a12} are less accurate when rating responses in low-resource varieties such as DA.}

\paragraph{Quality}

We next measure LLMs' fluency and semantic accuracy in DA, first by English-to-DA MT. 
Once more, we treat English as a proxy for the LLM's semantic knowledge; this measures how well the LLM can express a variety of semantic concepts in DA. 
We supplement automatic MT scores with human evaluations of adequacy and fluency of translations, as well as dialectness judgments, detailed in \S \ref{sec:hum_eval}.
We couple this with correlative evaluation to reveal whether the LLM's output quality deteriorates in DA compared to MSA. 
For this we gather all human DA fluency annotations previously collected and correlate them with dialectness scores, to determine whether fluency degrades in more dialectal generation. (See \S \ref{sec:hum_eval}.) 

\paragraph{Diglossia}

To measure LLMs' diglossic proficiency, we evaluate MSA$\leftrightarrow$DA MT. Table \ref{tab:competencies} summarizes details of all four evaluation competencies.

\subsection{Evaluation Corpora}
\label{sec:corpus}

In addition to NADI and ALDi models, we require three specialized corpora for evaluation: (1) \textbf{Cross-lingual} prompts, or English user inputs explicitly requesting specific DA varieties; (2) \textbf{Monolingual} prompts, or user requests in various DA varieties; and (3) \textbf{Bitext} prompts, or aligned bitexts with translations in English, MSA, and DA varieties. 



For the \textbf{cross-lingual} corpus, we adapted a set from \citet{marchisio2024understanding} of English LLM prompts with explicit requests for responses in different languages, and substituted the names of DA varieties. 
This set originally came from three distinct collections of LLM user prompts: a subset of Okapi \cite{lai-etal-2023-okapi} inputs to the Alpaca LLM \cite{taori2023alpaca}, a collection of ChatGPT inputs scraped from the ShareGPT API, and a corpus of human-curated prompts commissioned by \citeauthor{marchisio2024understanding} (denoted \textit{Cohere}). 
For the \textbf{monolingual} and \textbf{bitext} corpora, meanwhile, we selected four existing DA data sets based on their style diversity and dialectal coverage. 

We used two multi-variety DA bitext corpora, integrating 200 sentence pairs from each in our \textbf{bitext} corpus and 100 sentences from the monolingual DA portion of each in our \textbf{monolingual} corpus. 
The first, MADAR-26 \cite{bouamor-etal-2018-madar}, is a multi-way parallel bitext with English, MSA, and DA from 25 Arab League cities. 
The English source sentences were sourced from the Basic Traveling Expression Corpus (BTEC) \cite{takezawa-etal-2007-multilingual} and manually translated into the 26 Arabic varieties.\footnote{We used sets for Riyadh, Damascus, Jerusalem, Khartoum, Cairo, Algiers, and Fes to render seven countries' DA.} 
The genre of this corpus is BTEC, i.e. everyday utterances that might be expressed verbally. 
The second set, FLORES-200 \cite{team2022language}, is an MT evaluation benchmark 
of 1012 sentences in 204 language varieties. 
The English source texts were sampled from wiki sites and then translated manually. 
We use the sets for Najdi, North Levantine, South Levantine, Egyptian, and Moroccan Arabic to represent KSA, Syria, Palestine, Egypt, and Morocco, respectively. 

We then used two multi-variety monotext DA corpora, adding 100 additional sentences per variety from each to our \textbf{monolingual} corpus. 
The first, MADAR-Twitter \cite{bouamor-etal-2019-madar}, contains 2,980 tweets from 4,375 profiles in 21 Arab countries. 
The tweets were seeded from 25 hashtags representing Arab League states (e.g. \#Kuwait, \#Egypt) and labeled by country. 
The second corpus, HABIBI \cite{el-haj-2020-habibi}, contains $\>$30k song lyrics by artists from 18 Arab countries.\footnote{A native speaker of two DA varieties manually cleaned mislabeled sentences from the HABIBI sets we used, since some artists wrote in a DA variety other than their own. This annotator also deliberately selected lyrics for our 100-sentence subsets to highlight distinctions between varieties.} 
We used eight country-level subsets from each of these collections (corresponding to all eight countries in Figure \ref{fig:dial_reg}). 

Note that the purpose of \textbf{monolingual} evaluation is to measure how well an LLM matches a user's input DA variety, so it should be composed of prompts instruction-tuned LLMs are accustomed to. 
The data sources we used for monolingual sentences provided generic sentences of four genres: BTEC, wiki text, tweets, and song lyrics. 
To transform these diverse generic sentences into instructions, we used eight instruction templates (shown in Table \ref{tab:templates}). 
We surveyed native speakers of the eight DA varieties covered to retrieve translations of the templates. 
Then for each generic sentence, we randomly selected one of the templates in the appropriate variety and inserted the generic sentence, transforming it into a command. 
Also randomly chosen along with the template was the sentence location: i.e. whether it was inserted at the start or end of the template, or in the middle where applicable (see Table \ref{tab:templates}).



\begin{figure}
    \centering
    \includegraphics[width=.5\linewidth]{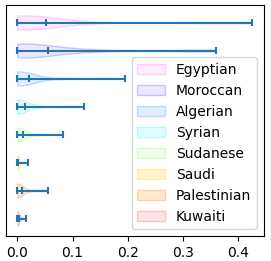}
    \caption{ADI2 (correct-variety dialectness scores) distributions across LLMs and genres in the crosslingual task (which requests specific DA varieties of the LLM in English). ADI2=0 indicates the wrong Arabic variety.} 
    \label{fig:xling-res}
\end{figure} 

\section{Evaluation and Results}

We detail our evaluation of nine LLMs for eight country-level DA varieties:\footnote{listed roughly from east to west} Kuwait, Saudi Arabia, Syria, Palestine, Sudan, Egypt, Algeria, and Morocco. 
These constitute two varieties from each of four Arabic dialectal regions (see Figure \ref{fig:dial_reg}).

\subsection{Primary Results}
\label{sec:prim_res}

\begin{figure*}
    \centering
    \includegraphics[width=\linewidth]{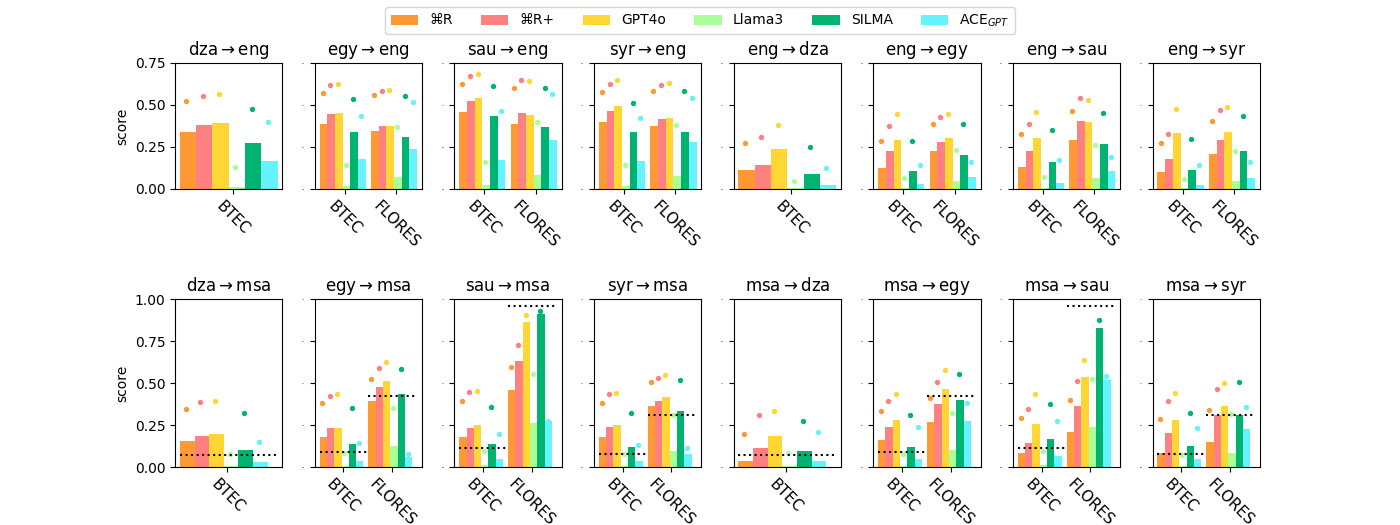}
    \caption{
    DA$\rightarrow$Eng MT surpasses Eng$\rightarrow$DA. DA$\leftrightarrow$MSA scores are low in the BTEC genre and rarely above the dotted-line zero-translate SpBLEU baseline for FLORES. Bars represent SpBLEU, while marks are chrF. Scores are between 0 and 1. (i.e. 0.5 corresponds to 50 SpBLEU points.) 
    Note \texttt{dza} is the country code for Algeria. 
    }
    \label{fig:mt-res}
\end{figure*}

Recall from \S \ref{sec:meth} that we evaluate \textbf{Fidelity}, via ADI2 score and macro-score; and \textbf{Understanding}, \textbf{Quality}, and \textbf{Diglossia}, via DA$\leftrightarrow$English and DA$\leftrightarrow$MSA MT. 
We begin with \textbf{Fidelity} and present scores across eight dialects, four genres, and nine LLMs including six open-source LLMs Command-R (and base model),\footnote{\url{https://cohere.com/blog/command-r}} Command-R+ (and base model),\footnote{\url{https://cohere.com/blog/command-r-plus-microsoft-azure}} and Llama 3.1 (and base model) \cite{dubey2024llama3herdmodels}; 
one closed LLM, GPT-4o;\footnote{\url{https://openai.com/index/hello-gpt-4o/}}
and two LLMs specialized in Arabic, ACEGPT \cite{huang-etal-2024-acegpt} (selected because of its prominence) and SILMA \cite{silma_01_2024} (selected because it led the Arabic LLM leaderboard\footnote{\url{https://huggingface.co/spaces/OALL/Open-Arabic-LLM-Leaderboard}, as of 27-10-2024} during our evaluation).
Results for monolingual and cross-lingual prompts are in Figures \ref{fig:mono-res} and \ref{fig:xling-res}, respectively (more detailed results in Appendix \ref{sec:appendix}).

Regarding \textbf{Fidelity}, the order of model performance is roughly consistent across genres and dialects. 
In \textbf{monolingual} settings, the Command-R+ base model, Llama-3.1, and Llama-3.1 base model performed best,\footnote{impressively, given its 8B parameters (see Table \ref{tab:mod_sizes}).} followed by Command-R base. 
Command-R, Command-R+, GPT-4o, and ACEGPT typically perform worse, and SILMA worst of all (see Fig. \ref{fig:mono-res}). 
Almost all ADI2 scores fell below 50\%. 
The Command base models' higher performance supports Hypothesis \ref{hyp:sft_bad} that post-training can inhibit DA modeling. 
All LLMs performed poorly on the \textbf{cross-lingual} task. 
Base models, ACEGPT, and Llama 3.1 were unable to complete the cross-lingual task whatsoever, and the performance of Command-R, Command-R+, and SILMA was low. 
GPT-4o was the only model to surpass 13\% ADI2, and only on half of the dialects. 
We summarize results in Figure \ref{fig:xling-res}.\footnote{See Figure \ref{fig:xling-bars} for fuller results.}
Note that on the lowest performing DA varieties, no LLMs exceed 20\% ADI2 on any genre for either task, and that even on Egyptian and Moroccan, a majority of LLMs still score below this threshold. 

A majority of LLM responses fail because they are in MSA rather than DA: ALDi and ADI2 scores are highly correlated with $\rho=0.99$ for cross-lingual and $\rho=0.91$ for monolingual, indicating that if models do not respond in the right dialect, they are typically not responding in DA at all. 
Manual inspection of responses suggests that LLMs output pure MSA (with ALDi=0 or nearly 0) by default, while occasionally responding more dialectally. 
(Figure \ref{fig:hists} illustrates this distribution.) 
We could not find any obvious predictors of the LLMs' dialectness from manual inspection.
This led us to hypothesize that DA responses are not triggered by any features other than random sampling:

\begin{hypothesis} \label{hyp:random}
    An LLM's distribution of correct DA responses does not correlate strongly with any detectable features of input text. 
\end{hypothesis}


Our MT automatic metric scores for four dialects are in Figure \ref{fig:mt-res}. 
(See Figure \ref{fig:mt-res-all} for the rest.)
GPT-4o and Command-R+ tended to perform best, with Command-R and SILMA usually lagging behind. 
ACEGPT typically performed worse and Llama 3.1 worse still. 
We forewent base models in this setting since MT is an instruction-oriented task. 
Model differences are more pronounced in directions evaluating \textbf{Diglossia} (DA$\leftrightarrow$MSA) and less so in those evaluating \textbf{Quality} and \textbf{Understanding}. 
Note that DA$\rightarrow$English MT scores are higher than English$\rightarrow$DA (supporting Hypothesis \ref{hyp:understanding} that LLM DA understanding beats generation). 
MSA$\leftrightarrow$DA MT is poor for BTEC. 
Dotted lines indicate zero-translation baseline scores for MSA$\leftrightarrow$DA (i.e. the SpBLEU score between source and target corpora without translation). 
In many cases the LLMs actually pull the source farther from the target, even when scores may appear high (such as for MSA$\leftrightarrow$Saudi Arabic FLORES). 
Overall, LLM DA$\leftrightarrow$MSA performance is either low, below the zero-translation threshold, or barely above it. 
We conclude that \textbf{LLMs are not strongly diglossic.} 

\subsection{Human evaluation}
\label{sec:hum_eval}



\begin{figure*}
    \centering
    \includegraphics[width=.49\linewidth]{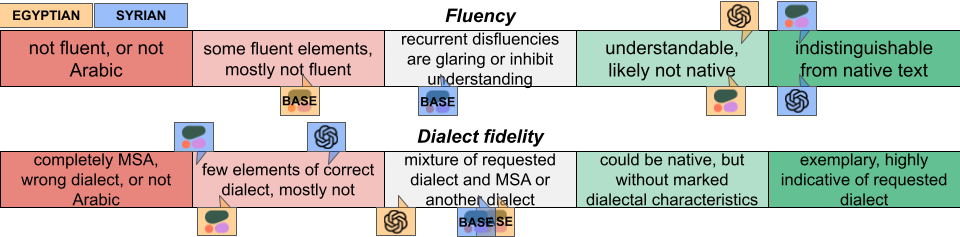}
    \includegraphics[width=.49\linewidth]{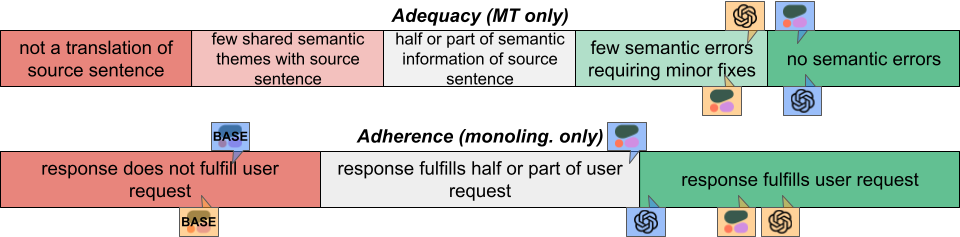}
    \caption{Human eval results shows that post-trained LLMs produce responses that are fluent, adequate, and adherent, but mostly not in the right DA variety. Command-R+ base improves dialectal fidelity but scores poorly on other metrics. Post-trained LLM fluency and fidelity scores were averaged across MT and monolingual tasks.}
    \label{fig:humeval_scale}
\end{figure*}

We asked two native speakers of Egyptian and one of Syrian Arabic to make judgments of the Command-R+ and GPT-4o DA outputs for both MT and monolingual tasks, as well as Command-R+ base for the monolingual task. 
Annotators judged \textit{Adherence} of monolingual responses, or how well the LLM fulfilled the user’s request, and \textit{Adequacy} of translations, or how well a translation reflected the meaning of the original sentence. 
For both tasks they made judgments of Arabic \textit{Fluency} and \textit{Dialectal Fidelity}, the final metric being the only one associated with the LLM’s ability to use the right DA variety. 
We defined the scales and guidelines for each of these measurements as shown in Figure \ref{fig:humeval_scale} and Table \ref{tab:humeval}. 
We averaged human scores across 50 prompts for each model. 
Annotators received prompts and completions in random order, for an unbiased review. 


See Figure \ref{fig:humeval_scale} for average human eval scores. 
Results indicate that the instruction fine-tuned models excel at (1) producing fluent Arabic text, (2) translating into Arabic with semantic adequacy, and (3) fulfilling DA requests, but that they struggle to do so in the correct DA variety. 
Their high \textit{adequacy} and low \textit{dialectal fidelity} scores (along with low ADI2 scores and high DA$\rightarrow$Eng MT scores in \S \ref{sec:prim_res}) indicate LLMs' \textbf{stronger DA understanding than generation}, supporting Hypothesis \ref{hyp:understanding}. This is an apparent reversal of the Generative AI Paradox \cite{west2024generative}, which highlights that LLMs' generative capabilities exceed their ability to understand their outputs. In contrast, we observe that LLMs are capable of understanding DA utterances but unable to produce fluent DA outputs.

Comparing base and post-trained models, Command-R+ base demonstrates better \textit{dialectal fidelity}, in part due to prompt reduplication, but fulfills user requests poorly with low fluency. We thus conclude that \textbf{post-training can harm DA proficiency, but is needed for other aspects of text quality} such as instruction following ability.

We correlated human fluency scores with dialectal fidelity to ascertain whether LLMs model DA with diminished fluency. 
However not a single one displayed significant negative correlation. 
The only relationship with $p<0.2$ was a weakly positive correlation of $\rho=.49$ for the Command-R+ base model in Egyptian. 
We thus conclude that \textbf{LLMs can produce DA with no perceptible decline in fluency} (when they are able to at all). 




\section{Follow-up Experiments and Analysis}
\label{sec:ana}



After conducting our primary evaluations, which produced ADI2 scores and both automatic and manual MT scores for each LLM, we analyzed LLM outputs further. 
Our motivation for this was (1) to explore inexpensive remedies to LLMs' DA deficiencies, and (2) to test Hypothesis \ref{hyp:random} and explore how user input features elicit LLM behaviors. 
We restrict this analysis to Command-R+, GPT-4o, and Llama 3.1 for depth and focus. 

\begin{figure}
    \centering
    \includegraphics[width=\linewidth]{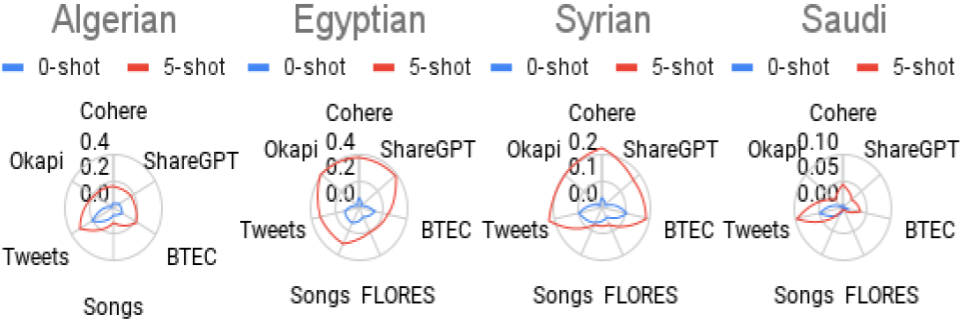}
    \caption{Command-R+ ADI2 scores from 5-shot prompting (red outer contours) are always higher than those of 0-shot prompting (blue inner contours).}
    \label{fig:fewshot}
\end{figure}

We first explored few-shot learning---an inexpensive mitigation approach---to improve DA modeling of the least performative of these LLMs: Command-R+. 
We curated five in-context prompt-completion examples for each of for DA varieties, for both monolingual and cross-lingual tasks, by translating the few shot examples used by \cite{marchisio2024understanding} in their related experiments.
Figure \ref{fig:fewshot} shows that \textbf{few-shot examples improve ADI2 across tasks and genres} for Command-R+. 

\begin{figure}
    \centering
    \includegraphics[width=1\linewidth]{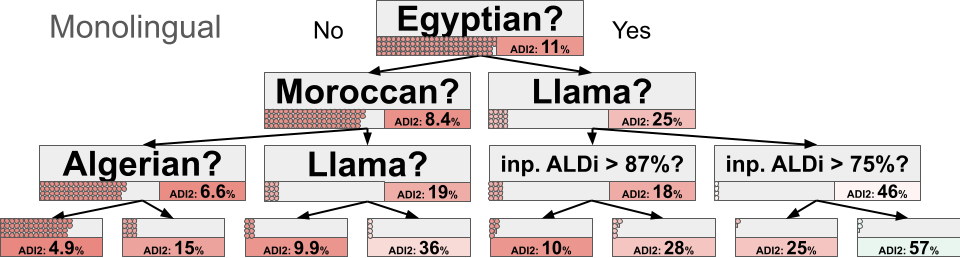} \\
    \includegraphics[width=1\linewidth]{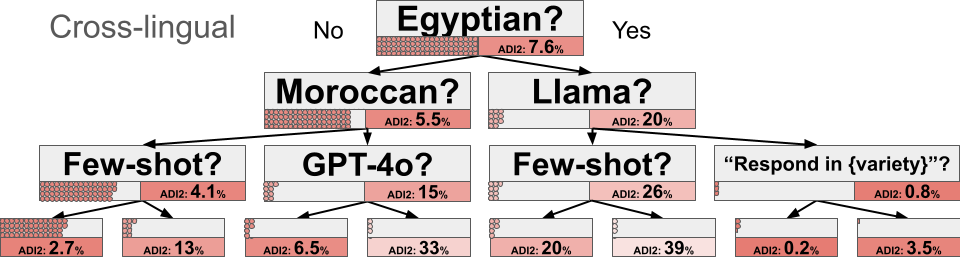}
    \caption{Decision trees regressed on ADI2 show that DA variety and LLM choice are primary dichotomous features. Each dot represents 100 samples, and color corresponds to group ADI2.}
    \label{fig:dec_trees}
\end{figure}

We then analyzed relationships between prompt attributes and performance. 
We collected features for each prompt and completion of the non-translation tasks: target DA variety; prompt length; prompt template; genre or data source; LLM used; number of few-shot examples; dialectness of input (for monolingual setting); and location of the dialect request (for cross-lingual) or inserted generic utterance (for monolingual) in the prompt. 
We fit decision tree regressors on ADI2 score using these features and \texttt{max\_depth} 3, shown in Figure \ref{fig:dec_trees}. 
The predominant dichotomous features are the desired DA variety and LLM used. 
Others include number of few-shot examples and dialectness of input text. 

For feature-level correlations with ADI2 score, we performed Spearman's rank tests for numerical features and ANOVA tests for categorical features to find $\rho$ and $\eta^2$ coefficients. 
We display these  only for the relationships where $p\<.001$, in Table \ref{tab:corr_coeff}, for both monolingual and cross-lingual tasks. 
No values in the table exceed \citeposs{Adams2014} threshold of strong correlation, $\eta^2\geq.14$, and no $\rho$ values exceed magnitude $0.5$. 
It seems target DA variety, LLM used, and number of few-shot examples correlate moderately with ADI2; but other features have weak or no correlation. 
In the monolingual task, the dialectness of the prompt also correlates significantly but weakly with both ADI2 and output dialectness ($\rho=.29$ and $\rho=.31$, respectively).\footnote{High input dialectness can result in a range of output dialectness, but low input dialectness typically precludes high output dialectness. See Fig. \ref{fig:aldi_scatter}.} 
\textbf{For a given LLM in a given DA variety, no feature of input text correlates strongly with DA performance.} 


\begin{table}[]
    \centering
    \small 
    \begin{tabular}{rccccc|cc}
         & \rotatebox{90}{DAV} & \rotatebox{90}{PT} & \rotatebox{90}{LOC} & \rotatebox{90}{GEN} & \rotatebox{90}{LLM} & \rotatebox{90}{LEN} & \rotatebox{90}{N} \\
         & \multicolumn{5}{c|}{$\eta^2=$} & \multicolumn{2}{c}{$\rho=$} \\
         \textbf{mono.} & .115 & .017 & \cellcolor[HTML]{d3d3d3} & .012 & .038 & -.09 & .12 \\
         \textbf{cross.} & .131 & \cellcolor[HTML]{d3d3d3} & .003 & \cellcolor[HTML]{d3d3d3} & .067 & \cellcolor[HTML]{d3d3d3} & .32 \\
    \end{tabular}
    \caption{Significant ($p\<.001$) correlations with ADI2. DAV=DA variety; PT=prompt template; LOC=DA request or generic utterance location; GEN=genre; LEN=prompt length; N=few-shot examples}
    \label{tab:corr_coeff}
\end{table}


\section{Conclusion}

We provide a comprehensive evaluation suite to measure LLM DA proficiency: Analyzing LLM Quality and Accuracy Systematically in Dialectal Arabic, or AL-QASIDA ("the poem" in Arabic).
We find that LLMs struggle to model DA, not due to faults in understanding or generation quality, but to a preference for MSA generation. 
Though more balanced pre-training data could likely mitigate this, we find that post-training can bias LLMs towards MSA, suggesting balanced post-training data as a lower-cost alternative. 
We also find few-shot examples can mitigate DA pitfalls at even lower cost. 
In the absence of mitigation strategies, we recommend GPT-4o for cross-lingual DA requests of Egyptian or Moroccan, and Llama-3.1 for monolingual DA generation. 
(Base models also do relatively well but have serious fluency liabilities.)




\section*{Limitations} 

In this work we did not explore all the DA modeling mitigation strategies we had originally intended to. 
(We decided to leave these for future work when the number of our results started to become unmanageably large.) 
Notably, we found promising preliminary indications that strategic preambles can help LLMs model DA better. 

Imaginably, we were unable to evaluate all popular LLMs and had to settle for relatively small a subset. 
We acknowledge this and emphasize that our development of this evaluation suite is meant as a technique for others to apply to other and future LLMs going forward. 
We did our best to include a diversity of LLMs so that readers may extrapolate general trends of LLM DA proficiency, but we acknowledge our sample may not be entirely representative. 

We also acknowledge that some of these LLMs, particularly the closed-source GPT-4o, are continuously updated and may not remain the same, even though the process of writing and publishing this paper. 
Hence, results drawn for GPT-4o may be slightly different as time passes and may not align perfectly with our findings here. 

\section*{Ethics Statement} 

DA modeling capabilities have a number of ethical implications. 
As we touched on in \S \ref{sec:intro}, LLMs have the potential both to create opportunities for the less advantaged, and to exacerbate existing inequalities. 
If LLMs are primarily proficient in MSA, as they are today, they may afford benefits only to Arabic speakers with enough education and social advantage to communicate comfortably in MSA, while those with less MSA proficiency may be left behind. 

The notion to treat Arabic as a monolith without representing its various diverse language varieties is often a result of homogeneity in the research community with a bias towards Western languages and values. 
We hope this work may bring a specific Arabic technological need to the community's attention as a small way to address this imbalance. 

We cover a diversity of eight DA language varieties by our evaluation, however we cannot claim to represent the varieties of all Arabic speakers, even in the eight countries represented. 
We hope our evaluation may be expanded to be more representative in the future. 

Lastly, we acknowledge that while all other evaluation sets we used were intended for this type of NLP evaluation, the song lyrics corpus we employed may have originally been intended for other purposes. 
Because most of these data sets were already curated for our general purposes, we did not vet them extensively for offensive content or sensative material. 
Our release of the song lyrics will not include any metadata (such as artist names), though such data can be found in its original source. 
Finally, we acknowledge use of LLMs to assist software development in this project, particularly in creating the graphics displayed in this paper. 

\bibliography{anthology,custom}

\appendix

\section{Supplemental Figures and Graphics}
\label{sec:appendix}

Here we present some supplemental visualizations. 
Table \ref{tab:gpt_convo} contains a conversation in which the user attempted to converse with GPT-4o in Egyptian DA, while the model repeated insisted on using MSA. 
Cell coloring corresponds to each utterance's dialectness per \citet{keleg-etal-2023-aldi}, listed in percentage form in the right column. 
Though the LLM's response on the sixth turn of conversation was partially DA (64\% dialectness overall), it did not satisfy the user because of its composition: the first three words of the utterance were 93\% dialectal, after which the LLM abruptly switched back into MSA (9\% dialectal) for the rest. 
Not until the fourteenth turn of conversation was the LLM's response satisfactorily dialectal, at 75\%. 
By comparison, the highest average dialectness score that any model achieved in our full evaluation presented here was only 61.4\% (reached by Llama-3.1 on Egyptian Arabic tweets in the monolingual task). 
The original Arabic conversation is likewise displayed in Table \ref{tab:gpt_convo_3araby}.

Table \ref{tab:templates} shows English versions of the templates we used to transform generic DA sentences into DA instructions (i.e. user inputs for an LLM). 
Native Arabic speakers from the eight countries indicated in Figure \ref{fig:dial_reg} translated the templates into their native DA varieties for our evaluation suite. 
Generic monolingual sentences from monotext or bitext corpora were inserted at different positions in the templates stochastically, see Table \ref{tab:template_locs}.

\begin{figure*}
    \centering
    \includegraphics[width=\linewidth]{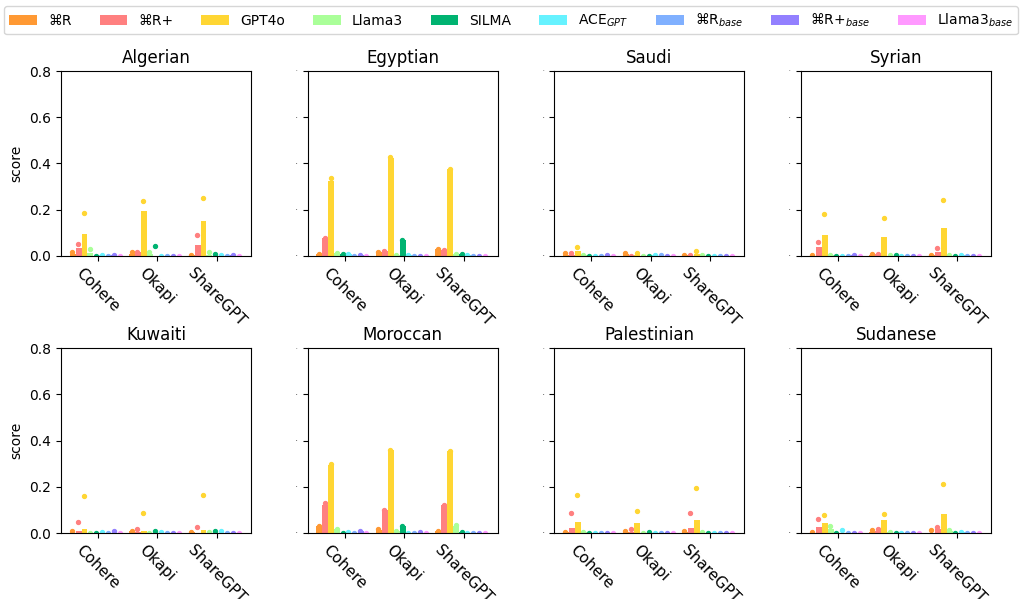}
    \caption{DA fidelity across genres, models, and dialects with \textbf{cross-lingual prompts}. Bars indicate ADI2 scores, while marks are macro-scores.}
    \label{fig:xling-bars}
\end{figure*}

\begin{figure*}
    \centering
    \includegraphics[width=1\linewidth]{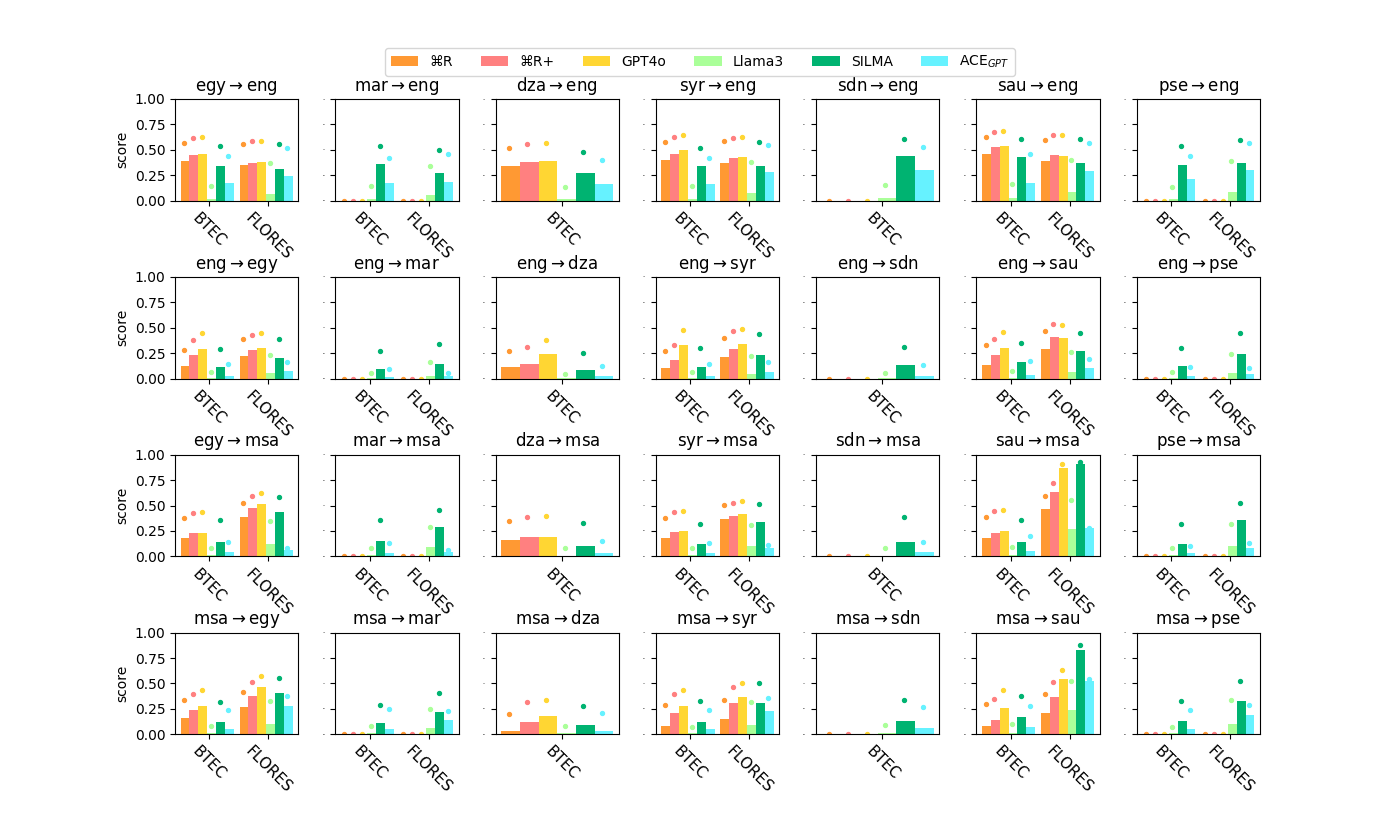}
    \caption{MT scores for all eight dialects. Compare to Figure \ref{fig:mt-res}. Bars represent SpBLEU, while marks are chrF.}
    \label{fig:mt-res-all}
\end{figure*}

Supplemental and more complete experimental results can be found in Figures \ref{fig:xling-bars} and \ref{fig:mt-res-all}
Figure \ref{fig:mt-res-all} visualizes MT scores as in Figure \ref{fig:mt-res}, but for all eight dialects evaluated. 
We will host complete detailed evaluation results on a server upon formal publication of this work. 

\begin{figure}
    \centering
    \includegraphics[width=1\linewidth]{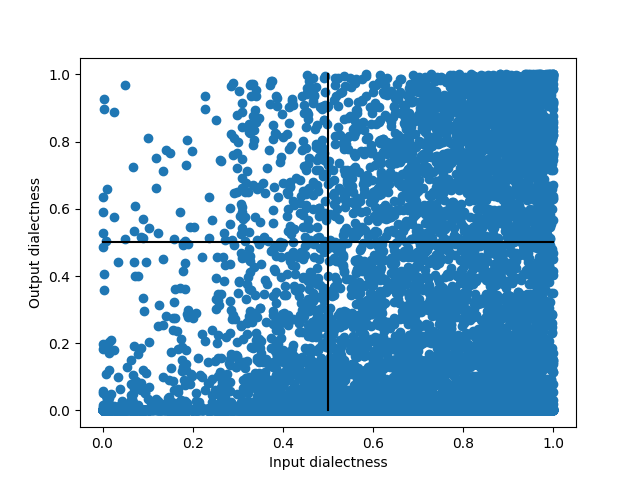}
    \caption{Prompt (input) and completion (output) dialectness values for monolingual evaluation; lines delineate quartiles
    } 
    \label{fig:aldi_scatter}
\end{figure}

Input dialectness is plotted against output dialectness in the monolingual task in Figure \ref{fig:aldi_scatter}. 
The correlation between these variables is positive, with Spearman's $\rho=.31$. 
Notice that high input dialectness can result in a diverse range of output dialectness values, but low input dialectness typically precludes high output dialectness. 
In other words, even high input dialectness does not guarantee much about the output dialectness, but low input dialectness more or less destroys any chances of dialectal output. 

We show the sizes of LLMs we used, in number of parameters, in Table \ref{tab:mod_sizes}. 
It is notable that Llama-3.1 scored best in the monolingual task, since we used only the 8B-parameter model, compared to much larger models Command-R+ and GPT-4o. 
We did not train any models and only performed inference. 

\begin{table}[]
    \centering
    \small 
    \begin{tabular}{rccr}
    
    \emoji{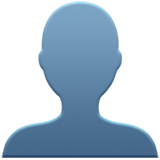} & \multicolumn{2}{R{.7\linewidth}}{\cellcolor[HTML]{62c092} What's up man, how are you?} & 97\% \\  
    \emoji{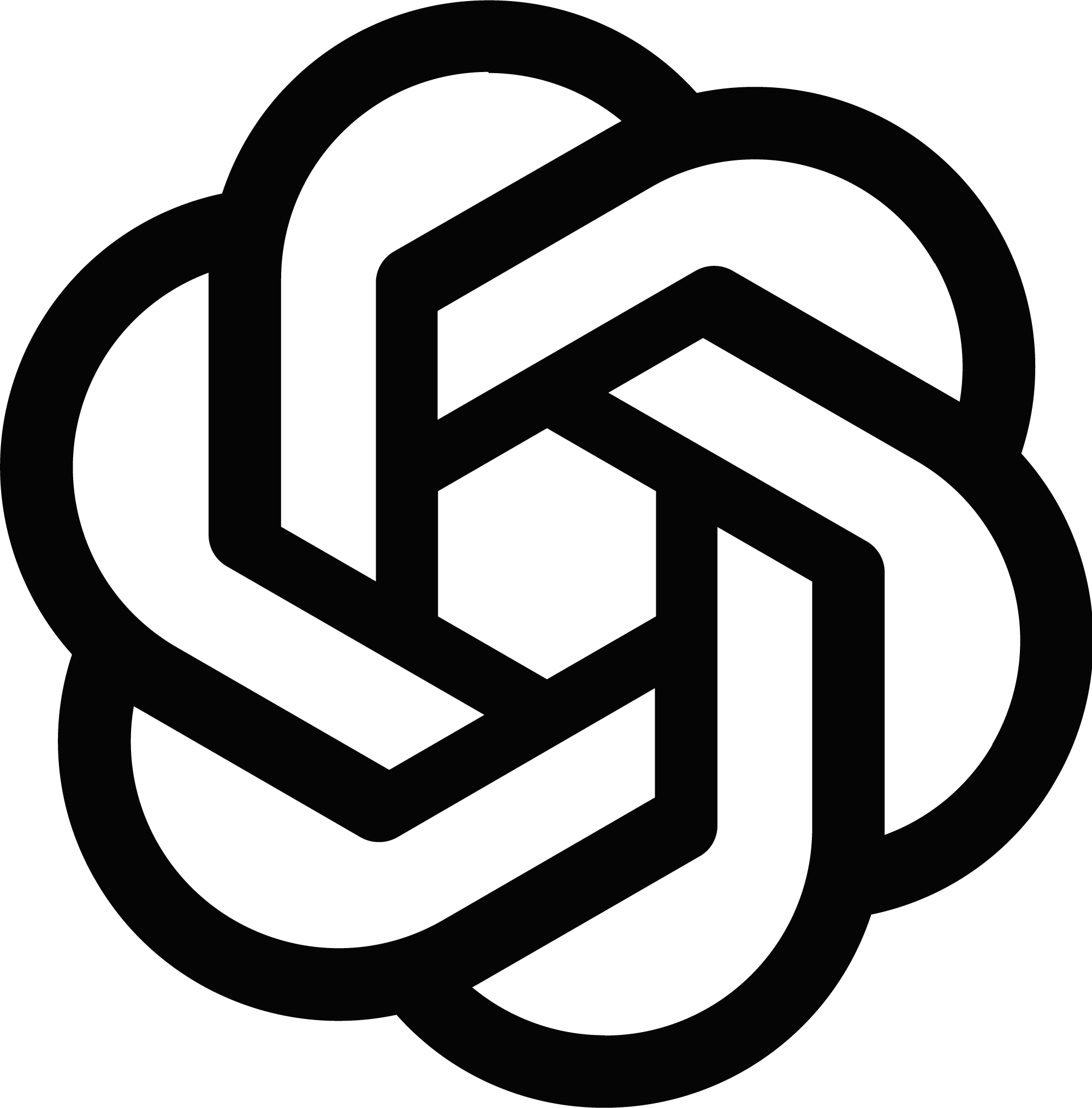} & \multicolumn{2}{p{.7\linewidth}}{\cellcolor[HTML]{efada8} I'm doing well, thank you for asking, and you? How are you?} & 19\% \\  

    \emoji{bust-in-silhouette_1f464.png} & \multicolumn{2}{R{.7\linewidth}}{\cellcolor[HTML]{6fc59b} All good, praise God. What are you up to today?} & 93\% \\  
    \emoji{openai_logo.png} & \multicolumn{2}{p{.7\linewidth}}{\cellcolor[HTML]{e67c73} I am engaging in assistance here via answering people's questions, presenting information, and helping to solve problems. How may I assist you today?} & 0\% \\  

    \emoji{bust-in-silhouette_1f464.png} & \multicolumn{2}{R{.7\linewidth}}{\cellcolor[HTML]{80cca7} My goodness, very awesome! But why are you speaking MSA? Talk to me a bit in Egyptian.} & 88\% \\  
    \emoji{openai_logo.png} & \multicolumn{2}{p{.7\linewidth}}{\cellcolor[HTML]{d0ecdf} Sure, no problem. If you don't mind, is there anything specific you require assistance with today?} & 64\% \\

    \emoji{bust-in-silhouette_1f464.png} & \multicolumn{2}{R{.7\linewidth}}{\cellcolor[HTML]{5ebe8f} Yeah I just wanted a partner to speak together in the Egyptian dialect, the best Arabic dialect!} & 98\% \\  
    \emoji{openai_logo.png} & \multicolumn{2}{p{.7\linewidth}}{\cellcolor[HTML]{f5cac7} Sure, I'm here to help you. That's a very beautiful thing, because I love communicating in the different dialects. If you have any question or need help with anything, just ask.} & 30\% \\

    \emoji{bust-in-silhouette_1f464.png} & \multicolumn{2}{R{.7\linewidth}}{\cellcolor[HTML]{6fc59b} Great. But you're still speaking MSA a lot. I appreciate your efforts, but maybe you're not very skilled in the Egyptian dialect?} & 93\% \\  
    \emoji{openai_logo.png} & \multicolumn{2}{p{.7\linewidth}}{\cellcolor[HTML]{e67c73} I am here to help in any language you like, and naturally I try to the best of my abilities to speak in the prefered dialect of the interlocutor. But if you desire to communicate in the Egyptian dialect, then let us begin our conversation in the Egyptian dialect now. Is there something you would like to discuss or something to help you with?} & 0\% \\

    \emoji{bust-in-silhouette_1f464.png} & \multicolumn{2}{R{.7\linewidth}}{\cellcolor[HTML]{8dd1b0} Sure, just tell me about your interests now. What do you like to do?} & 84\% \\  
    \emoji{openai_logo.png} & \multicolumn{2}{p{.7\linewidth}}{\cellcolor[HTML]{e67c73} Certainly. I love to help people and to present accurate and useful information. Aside from that, I also enjoy reading books and examining the latest updates in the fields of technology and science. And as you know, I am here to help you in any matter that you need help in, so do not hesitate to ask any question.} & 0\% \\

    \emoji{bust-in-silhouette_1f464.png} & \multicolumn{2}{R{.7\linewidth}}{\cellcolor[HTML]{97d5b7} You are still using totally MSA words. like "dhaalik" and "ayDan" and "aahkir". Talk to me in Egyptian from now on, boss. Try to tell me more about your interests and hobbies in the dialect please.} & 81\% \\  
    \emoji{openai_logo.png} & \multicolumn{2}{p{.7\linewidth}}{\cellcolor[HTML]{abddc5} You're right, man. Let's ditch the MSA and continue in Egyptian. I love to help people, and at the same time I like to read books and learn new things. For hobbies, I love to travel and to discover new places, and I also love nature and wandering in it. And in my free time I like to watch movies and listen to music. If there's anything you'd like to know or ask about, let me know, pal.} & 75\% \\
    
    
    \end{tabular}
    \caption{Arabic conversation with GPT-4o, translated into English. Percentages on the right represent dialectness scores from \citet{keleg-etal-2023-aldi}.
    }
    \label{tab:gpt_convo}
\end{table} 

\begin{table}[]
    \centering
    \tiny  
    \begin{tabular}{rrr}
    
    \emoji{bust-in-silhouette_1f464.png} & \cellcolor[HTML]{62c092} \ar{عامل ايه يا حبيبى, ازيك؟} & 97\% \\  
    
    \emoji{openai_logo.png} & \cellcolor[HTML]{efada8} \ar{أنا بخير, شكراً لسؤالك. وأنت؟ كيف حالك؟} & 19\% \\  

    \emoji{bust-in-silhouette_1f464.png} & \cellcolor[HTML]{6fc59b} \ar{كويس الحمد لله. عم تعمل ايه النهاردا بقى؟} & 93\% \\  
    
    \emoji{openai_logo.png} & \cellcolor[HTML]{e67c73} \ar{أقوم بالمساعدة هنا عبر الإجابة على أسئلة} & 0\% \\  
     & \cellcolor[HTML]{e67c73} \ar{الناس وتقديم المعلومات والمساعدة في حل} &  \\
     & \cellcolor[HTML]{e67c73} \ar{المشاكل. كيف يمكنني مساعدتك اليوم؟} &  \\

    \emoji{bust-in-silhouette_1f464.png} & \cellcolor[HTML]{80cca7} \ar{رائع اوى ما شاء الله. بس عم تتكلم فصحى} & 88\% \\  
     & \cellcolor[HTML]{80cca7} \ar{ليه؟ كلمنى مصرى شويه} &  \\  
    
    \emoji{openai_logo.png} & \cellcolor[HTML]{d0ecdf} \ar{طبعًا, مفيش مشكلة. بعد إذنك, هل يوجد} & 64\% \\
     & \cellcolor[HTML]{d0ecdf} \ar{أي شيء بالضبط تحتاج مساعدة فيه اليوم؟} &  \\

    \emoji{bust-in-silhouette_1f464.png} & \cellcolor[HTML]{5ebe8f} \ar{ايوا بس كنت عايز شريك عشان نتكلم} & 98\% \\  
     & \cellcolor[HTML]{5ebe8f} \ar{مع بعض باللهجه المصريه, احسن لهجه} &  \\  
     & \cellcolor[HTML]{5ebe8f} \ar{عربيه والله!} &  \\
    
    \emoji{openai_logo.png} & \cellcolor[HTML]{f5cac7} \ar{طبعًا, أنا هنا لمساعدتك. ده شيء جميل جدًا,} & 30\% \\
     & \cellcolor[HTML]{f5cac7} \ar{فأنا أحب التواصل باللهجات المختلفة. إذا} &  \\
     & \cellcolor[HTML]{f5cac7} \ar{كان لديك أي سؤال أو شيء تحتاج مساعدة} &  \\
     & \cellcolor[HTML]{f5cac7} \ar{فيه, ما عليك إلا أن تسأل.} &  \\

    \emoji{bust-in-silhouette_1f464.png} & \cellcolor[HTML]{6fc59b} \ar{تمام. بس لسه عم تتكلم فصحى كتير. بقدر} & 93\% \\ 
     & \cellcolor[HTML]{6fc59b} \ar{محاولاتك, بس ممكن انو انت مش ماهر} &  \\
     & \cellcolor[HTML]{6fc59b} \ar{اوى باللهجه المصريه؟} &  \\
    
    \emoji{openai_logo.png} & \cellcolor[HTML]{e67c73} \ar{أنا هنا للمساعدة بأي لغة تفضلها, وبالطبع } & 0\% \\
     & \cellcolor[HTML]{e67c73} \ar{أحاول قدر المستطاع التحدث باللهجة التي} &  \\
     & \cellcolor[HTML]{e67c73} \ar{يفضلها المحادث. لكن إذا كنت ترغب في} &  \\
     & \cellcolor[HTML]{e67c73} \ar{التواصل باللهجة المصرية, فلنبدأ محادثتنا} &  \\
     & \cellcolor[HTML]{e67c73} \ar{باللهجة المصرية من الآن. هل هناك شيء} &  \\
     & \cellcolor[HTML]{e67c73} \ar{ترغب في الحديث عنه أو مساعدتك فيه؟} &  \\

    \emoji{bust-in-silhouette_1f464.png} & \cellcolor[HTML]{8dd1b0} \ar{ايوا بس كلمنى عن اهتماماتك دالوقت.} & 84\% \\
     & \cellcolor[HTML]{8dd1b0} \ar{تحب تعمل ايه بقى؟} &  \\
    
    \emoji{openai_logo.png} & \cellcolor[HTML]{e67c73} \ar{طبعًا, بكل سرور. أنا أحب مساعدة الناس} & 0\% \\
     & \cellcolor[HTML]{e67c73} \ar{وتقديم المعلومات الصحيحة والمفيدة.} &  \\
     & \cellcolor[HTML]{e67c73} \ar{بجانب ذلك, أيضًا أستمتع بقراءة الكتب} &  \\
     & \cellcolor[HTML]{e67c73} \ar{والاطلاع على آخر التطورات في مجالات} &  \\
     & \cellcolor[HTML]{e67c73} \ar{التكنولوجيا والعلوم. وكما تعلم, أنا هنا} &  \\
     & \cellcolor[HTML]{e67c73} \ar{لمساعدتك بأي موضوع تحتاج مساعدة} &  \\
     & \cellcolor[HTML]{e67c73} \ar{فيه, فلا تتردد في طرح أي سؤال.} &  \\

    \emoji{bust-in-silhouette_1f464.png} & \cellcolor[HTML]{97d5b7} \ar{لسه عم تستعمل كلمات فصحى خالص} & 81\% \\  
     & \cellcolor[HTML]{97d5b7} \ar{زي "ذلك" و"أيضا" و"آخر". كلمنى} &  \\
     & \cellcolor[HTML]{97d5b7} \ar{مصرى من دالوقت يا باشا. حاول تخبرنى} &  \\
     & \cellcolor[HTML]{97d5b7} \ar{اكتر عن اهتماماتك وهوياتك باللهجه دى} &  \\
     & \cellcolor[HTML]{97d5b7} \ar{لو سمحت.} &  \\
    
    \emoji{openai_logo.png} & \cellcolor[HTML]{abddc5} \ar{صح يا حبيبي, هونسيب الفصحى ونكمل} & 75\% \\
     & \cellcolor[HTML]{abddc5} \ar{بالمصري. أنا بحب أساعد الناس, وفي نفس} &  \\
     & \cellcolor[HTML]{abddc5} \ar{الوقت بحب أقرأ كتب وأتعلم حاجات جديدة,} &  \\
     & \cellcolor[HTML]{abddc5} \ar{بخصوص الهوايات, بحب السفر} &  \\
     & \cellcolor[HTML]{abddc5} \ar{واستكشاف أماكن جديدة, وكمان بحب} &  \\
     & \cellcolor[HTML]{abddc5} \ar{الطبيعة والتجول فيها. وفي الوقت الفاضي} &  \\
     & \cellcolor[HTML]{abddc5} \ar{بحب أشوف أفلام وأسمع موسيقى. إذا في} &  \\
     & \cellcolor[HTML]{abddc5} \ar{حاجة تحب تعرفها أو تسأل عنها, قولي يا} &  \\
     & \cellcolor[HTML]{abddc5} \ar{زميل.} &  \\
    
    \end{tabular}
    \caption{Arabic conversation with GPT-4o, in original form. Percentages on the right represent dialectness scores from \citet{keleg-etal-2023-aldi}.
    }
    \label{tab:gpt_convo_3araby}
\end{table}

\begin{table}[]
    \centering
    \small
    \begin{tabular}{rp{.9\linewidth}}
         0 & Paraphrase [this] in your own words, preserving the original meaning. \\
         1 & Craft a creative response that incorporates this sentence as a key element. \\
         2 & Write a story beginning with this line. \\
         3 & What are the implications of [this], if someone said it to you? \\
         4 & Rewrite [this sentence] to be more appropriate for general audiences. \\
         5 & What audience is this type of sentence intended for? Justify your answer. \\ 
         6 & Reply to this thread. \\
         7 & Draft a potential exam question, in an essay format, based on [this sentence.] \\
         
    \end{tabular}
    \caption{Instruction templates to transform generic sentences into LLM commands. Brackets [] indicate the location of a sentence insertion, if inserted in the middle (rather than at the beginning or end).}
    \label{tab:templates}
\end{table}

\begin{table}[]
    \centering
    \small
    \begin{tabular}{rp{.7\linewidth}}
         beginning & [] Paraphrase this in your own words, preserving the original meaning. \\
         middle & Paraphrase [] in your own words, preserving the original meaning. \\
         end & Paraphrase this in your own words, preserving the original meaning: [] \\
         
    \end{tabular}
    \caption{Generic sentences may be inserted into any template at the beginning, middle, or end of a template (at any of the [] sites in this example). This variable was chosen stochastically for each generic sentence in creating the evaluation suite.}
    \label{tab:template_locs}
\end{table}

\begin{table}
    \centering
    \begin{tabular}{rl}
        Command-R & 35B \\
        Command-R+ & 104B \\
        Llama-3.1 & 8B \\
        SILMA & 9B \\
        ACEGPT & 7B \\
        GPT-4o & ?? (likely $>$175B) \\
    \end{tabular}
    \caption{Sizes as parameter counts of the models used in this study}
    \label{tab:mod_sizes}
\end{table}

\begin{table}[]
    \centering
    \begin{tabular}{rp{.9\linewidth}}
        \multicolumn{2}{c}{\textbf{Adherence}} \\
         3 & = the response fulfills user request completely \\ 
         2 & = the response fulfills half or part of the user request \\ 
         1 & = the response does not fulfill the user request at all \\ 
         \hline 
    \end{tabular}
    \begin{tabular}{rp{.9\linewidth}}
        \multicolumn{2}{c}{\textbf{Translation Adequacy}} \\
         5 &  = no semantic errors \\
         4 & = a few semantic errors that require minor fixes \\
         3 & = half or part of the semantic information of the source sentence preserved \\ 
         2 & = a few shared semantic themes with the source sentence \\ 
         1 & = not a translation of the source sentence \\ 
         \hline 
    \end{tabular}
    \begin{tabular}{rp{.9\linewidth}}
        \multicolumn{2}{c}{\textbf{Fluency}} \\
         5 &  = indistinguishable from native Arabic text \\
         4 & = understandable, but likely not native Arabic text \\
         3 & = clearly not native Arabic text, recurrent disfluencies are glaring or inhibit understanding (or includes copied text from the input prompt alongside newly generated text) \\ 
         2 & = some fluent elements, but mostly not fluent (or copies the input prompt without innovating) \\ 
         1 & = not fluent, or not Arabic \\ 
         \hline 
    \end{tabular}
    \begin{tabular}{rp{.9\linewidth}}
        \multicolumn{2}{c}{\textbf{Dialectal Fidelity}} \\
         5 &  = exemplary Arabic highly indicative of the requested dialect (could be native) \\
         4 & = could be native, but does not display exemplary or marked dialectal characteristics of the requested dialect (i.e. could be a number of dialects) \\
         3 & = a mixture of the requested dialect and MSA or another dialect \\ 
         2 & = a few elements of the correct dialect, but mostly not \\ 
         1 & = completely MSA, clearly the wrong dialect entirely, or not Arabic \\  
         * & Note dialectal fidelty is not necessarily penalized when the LLM copies the input prompt.
    \end{tabular}
    \caption{Human rating scales for different dimensions of output quality}
    \label{tab:humeval}
\end{table}

\begin{figure*}
    \centering
    \includegraphics[width=0.49\linewidth]{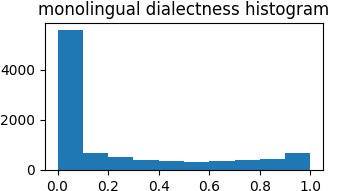}
    \includegraphics[width=0.49\linewidth]{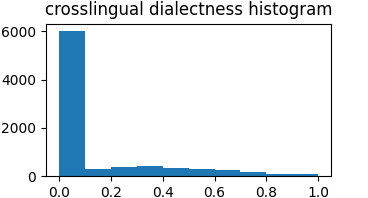}
    \caption{Distribution of ALDi (dialectness) scores. LLMs tend to produce standard Arabic (with ALDi$\approx$0) most of the time, while occasionally venturing into more dialectal responses.}
    \label{fig:hists}
\end{figure*} 

\end{document}